\documentclass[sigconf]{acmart}
\usepackage{multirow}
\usepackage{subfigure}
\usepackage[linesnumbered,lined,boxed,commentsnumbered]{algorithm2e}
\usepackage{color,xcolor}
\usepackage{pifont} 

\AtBeginDocument{%
  \providecommand\BibTeX{{%
    \normalfont B\kern-0.5em{\scshape i\kern-0.25em b}\kern-0.8em\TeX}}}

 \acmDOI{}

 \acmConference[ACM IH\&MMSEC'21]{%
   9th ACM Workshop on Information Hiding and Multimedia Security}{June 21--25, 2021}{IH\&MMSEC Brussels, Belgium}
 \acmPrice{}

\begin{document}

\title{Using contrastive learning to improve the performance of steganalysis schemes}

\author{Yanzhen Ren}
\email{renyz@whu.edu.cn}
\affiliation {%
  \institution{i. Key Laboratory of Aerospace Information Security and Trusted Computing, Ministry of Education.}
  \institution{ii. School of Cyber Science and Engineering, Wuhan University.}
  \country{China}
}

\author{Yiwen Liu}
\email{632918847@qq.com}
\affiliation{%
  \institution{i. Key Laboratory of Aerospace Information Security and Trusted Computing, Ministry of Education.}
  \institution{ii. School of Cyber Science and Engineering, Wuhan University.}
  \country{China}
}

\author{Lina Wang}
\email{lnawang@163.com}
\affiliation{%
  \institution{i. Key Laboratory of Aerospace Information Security and Trusted Computing, Ministry of Education.}
  \institution{ii. School of Cyber Science and Engineering, Wuhan University.}
  \country{China}
}
%


\begin{abstract}
  To improve the detection accuracy and generalization of steganalysis,
  this paper proposes the Steganalysis Contrastive Framework (SCF) based on contrastive learning.
  The SCF improves the feature representation of steganalysis by 
  maximizing the distance between features of samples of different categories
  and minimizing the distance between features of samples of the same category.
  To decrease the computing complexity of the contrastive loss in supervised learning,
  we design a novel Steganalysis Contrastive Loss (StegCL) 
  based on the equivalence and transitivity of similarity.
  The StegCL eliminates the redundant computing in the existing contrastive loss.
  The experimental results show that 
  the SCF improves the generalization and detection accuracy of existing steganalysis DNNs,
  and the maximum promotion is 2\% and 3\% respectively.
  Without decreasing the detection accuracy, 
  the training time of using the StegCL is 10\% of 
  that of using the contrastive loss in supervised learning.

\end{abstract}




\maketitle


\section{Introduction}

  Steganography generates the stego by embedding undetectable messages into the cover
  and makes them as close in distribution as possible \cite{wu2019novel, pevny2010using}.
  Steganalysis is to detect whether an object is a cover or a stego \cite{zhang2019depth, wang2018cnn, cogranne2019alaska}.

  According to the development history of steganalysis, steganalysis is divided into traditional steganalysis and DNN-based steganalysis.
  The traditional steganalysis scheme is to design handcrafted features followed by a classifier.
  The most classical traditional steganalysis scheme is the Rich Model \cite{fridrich2012rich} followed
  by an Ensemble Classifier \cite{kodovsky2011ensemble}.
  Recently, most steganalysis schemes are based on the DNN.
  Qian et al. \cite{qian2015deep} unify the feature extraction and classification two steps under a single DNN.
  XuNet \cite{xu2016structural} facilitates the hyperbolic tangent (TanH) activation function and $1\times1$ convolutional kernel.
  YeNet \cite{ye2017deep} utilizes the optimizable filters of Spatial Rich Model (SRM) \cite{fridrich2012rich} as the preprocessing module.
  Yedroudj et al. \cite{yedroudj2018yedroudj} introduce the batch normalization and non-linear activation function.
  Fridrich et al. \cite{boroumand2018deep} propose an end-to-end DNN called SRNet that introduces shortcut operation \cite{szegedy2017inception}.
  Zhao et al. \cite{you2020siamese} put forward a novel DNN with the siamese architecture.
  These DNN-based steganalysis schemes propose kinds of structures to optimize the DNN.
  The structures maximize the distance between features of cover and stego,
  which improves the detection accuracy of steganalysis.
  Taking the similarity between samples of the same category into account,
  this paper introduces contrastive learning to 
  minimize the distance between features of samples of the same category
  and maximize the distance between features of samples of different categories,
  which improves the detection accuracy and generalization of steganalysis DNNs.

  The main contributions of this paper can be summarized as follows:
  \begin{itemize}
    \item 
      To improve the detection accuracy and generalization of steganalysis,
      this paper proposes a Steganalysis Contrastive Framework (SCF) based on contrastive learning.
      The SCF improves the feature representation of steganalysis by 
      maximizing the distance between features of samples of different categories,
      and minimizing the distance between features of samples of the same category.
    \item 
      To decrease the computing complexity of supervised contrastive loss \cite{khosla2020supervised},
      we design a novel Steganalysis Contrastive Loss (StegCL) 
      based on the equivalence and transitivity of similarity.
      Without decreasing the detection accuracy,
      the StegCL eliminates the redundant computing 
      in the contrastive loss in supervised learning.
  \end{itemize}

  The rest of the paper is organized as follows: in Section II, the DNN-based steganalysis and the loss of 
  contrastive learning are introduced,
  in Section III, the SCF and the StegCL are explained,
  in Section IV, there are the setups of the experiments and the experimental results,
  and Section V summarizes the paper.

\section{Background}

\subsection{DNN-based steganalysis}

\begin{figure}[htbp]
  \centering
  \includegraphics[width=.97\linewidth]{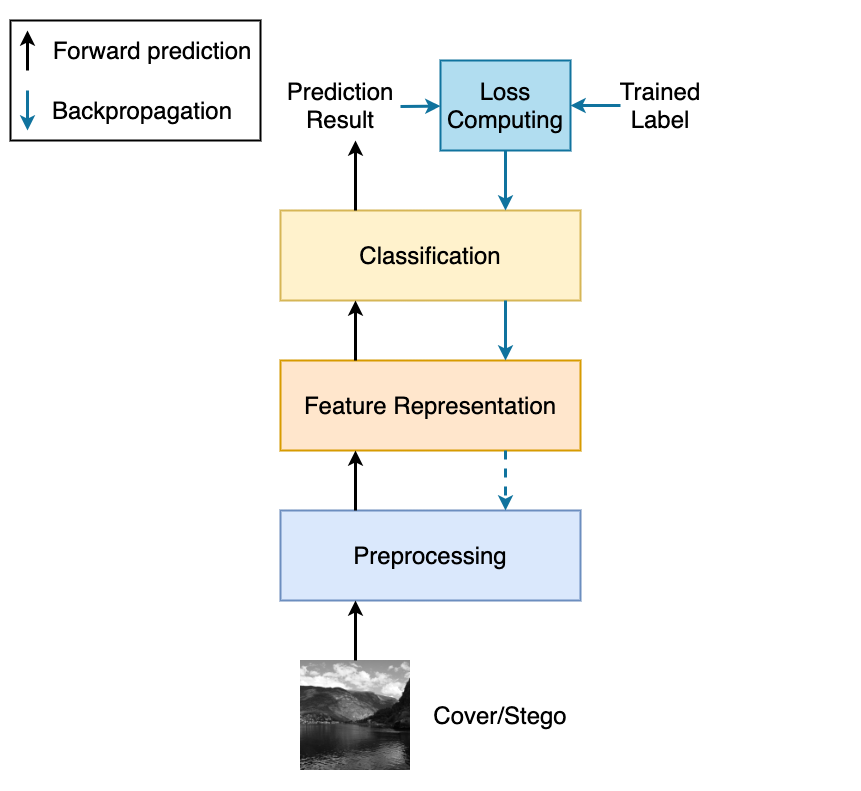}
  \caption{The basic framework of existing steganalysis DNNs.}
  \label{fig:2}
\end{figure}

  The basic framework of existing steganalysis DNNs is shown in Figure~\ref{fig:2}.
  There is the preprocessing module, feature representation module, and classification module in the framework.
  The preprocessing module filters the input image to obtain the high-frequency information of the image.
  The feature representation module extracts the high-dimensional features of the filtered image.
  The classification module classifies the high-dimensional features and obtains the prediction result.
  The black arrows indicate the process of forwarding prediction. During the forwarding prediction,
  the input image goes through preprocessing, feature representation, and classification modules in turn.
  The output of the forwarding prediction is the prediction result.
  The blue arrows indicate the process of backpropagation. During the backpropagation,
  the loss computing module computes the cross-entropy loss \cite{zhang2018generalized} with the prediction result and the trained label.
  Then, the computed loss is utilized by the backpropagation to update the parameters of the DNN.
  The dashed arrow between the feature representation module and the preprocessing module 
  means that sometimes
  the parameters of the preprocessing module are not updated \cite{xu2016structural}.

  Existing DNN-based steganalysis works focus on optimizing the preprocessing module and the feature representation module.
  To optimize the preprocessing module,
  Xu et al. \cite{xu2016structural} put forward to utilizing the fixed filters of SRM \cite{denemark2014selection}.
  Ye et al. \cite{ye2017deep} propose that the parameters of filters of SRM need to be updated.
  Fridrich et al. \cite{boroumand2018deep} utilize a convolution layer,
  a batch normalization layer, and a ReLU layer with default initialization as the preprocessing module.
  To optimize the feature representation module,
  Yedroudj et al. \cite{yedroudj2018yedroudj} design a non-linear activation function 
  called truncation function that maximizes the small difference between cover and stego. 
  SRNet \cite{boroumand2018deep} introduces the shortcut operation \cite{szegedy2017inception},
  which extracts more residual information between cover and stego.
  Zhao et al. \cite{you2020siamese} put forward KeNet with a novel siamese architecture
  according to the left and right imbalance of the difference between cover and stego.
  These works propose kinds of schemes to optimize the preprocessing module and the feature representation module.
  The schemes maximize the difference between features of cover and stego, which improves the detection accuracy of steganalysis.
  Inspired by contrastive learning that greatly improves the feature representation in CV and NLP \cite{chen2020simple, chen2020improved, jaiswal2021survey},
  this paper proposes the Steganalysis Contrastive Framework (SCF) based on contrastive learning
  to improve the performance of steganalysis schemes.

\subsection{The loss of contrastive learning}

\begin{figure}[htbp]
  \centering
  \includegraphics[width=0.83\linewidth]{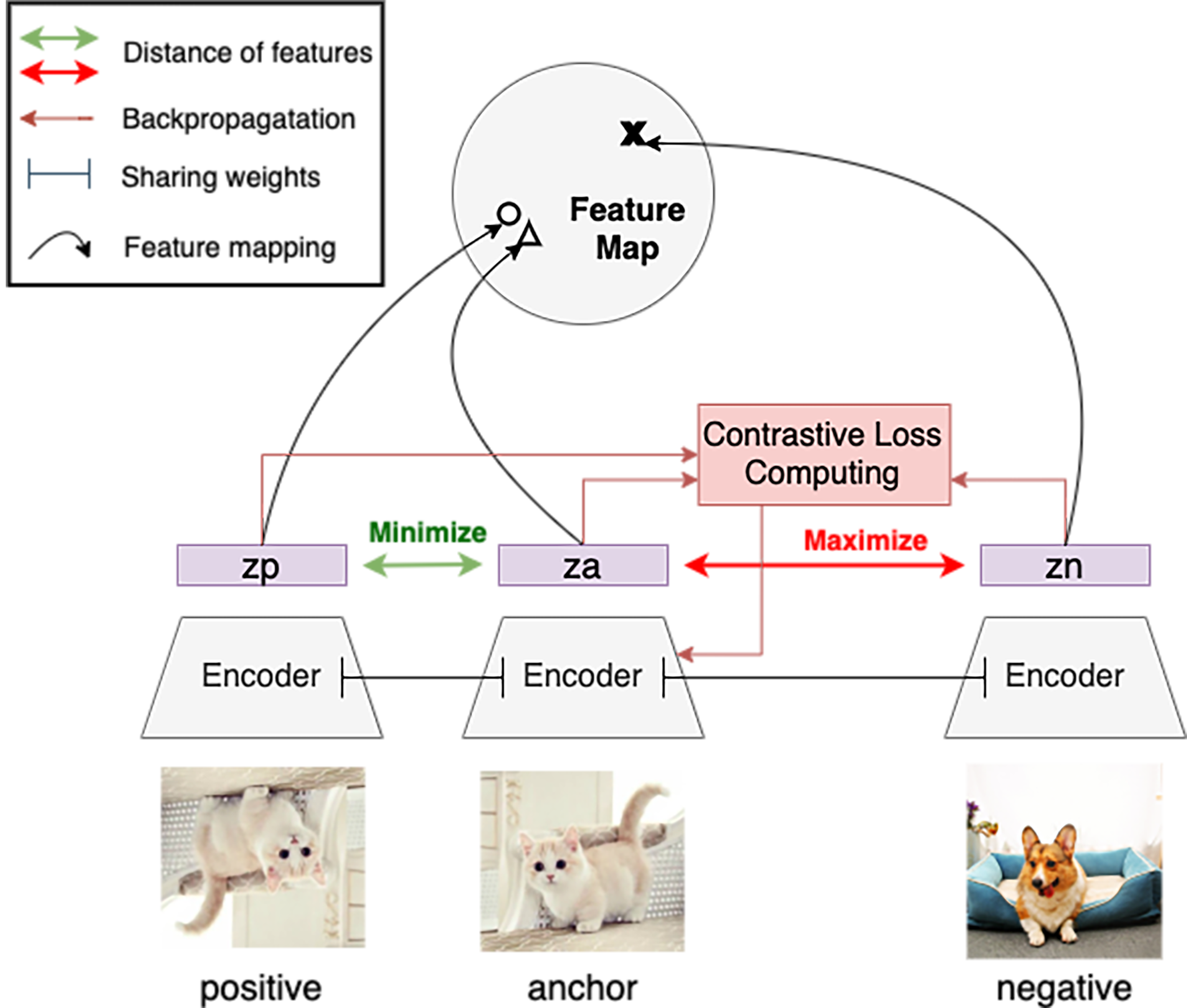}
  \caption{The basic principle of contrastive learning.}
  \label{fig:3}
\end{figure}

  The target of contrastive learning is to minimize the distance between features of the anchor and the positive, 
  and maximize the distance between features of the anchor and the negative.
  The basic principle of contrastive learning is shown in Figure~\ref{fig:3}.
  The three Encoders have the same structures and parameters.
  The $za$, $zp$, and $zn$ 
  are the features extracted from the anchor, the positive, and the negative by the Encoder.
  The contrastive loss computing module computes the contrastive loss with the 
  $za$, $zp$, and $zn$.
  The contrastive loss used by the backpropagation is the core of contrastive learning.
  As the backpropagation updates the parameters of the Encoder to decrease the contrastive loss,
  the $za$ and $zp$ gradually approach,
  and the $za$ and $zn$ gradually move away from each other, which achieves the target of contrastive learning.
  There are mainly two kinds of contrastive loss: Self-supervised Contrastive Loss (SelfCL) \cite{chen2020simple, chen2020improved, jaiswal2021survey} and 
  Supervised Contrastive Loss (SupCL) \cite{khosla2020supervised}.

\begin{equation}
	\mathbb{L}^{self}
	=
	\sum^{2N}_{i=1}\mathbb{L}^{self}_i
\end{equation}
\begin{equation}
	\mathbb{L}^{self}_{i}=-
	\log{
		\frac{\exp(z_i \cdot z_{j(i)} / \tau)}
		{\sum^{2N}_{k=1}\ell_{i \neq k} \cdot \exp(z_i \cdot z_k  / \tau)}
	}
\end{equation}

  The SelfCL is used in self-supervised learning \cite{jaiswal2021survey, chen2020simple, chen2020improved},
  In self-supervised learning, because of the lack of the labels of samples, the contrast occurs
  between the anchor, the augmentation of the anchor (the positive),
  and the samples different from the anchor (the negative).
  The SelfCL aims at maximizing the distance between features of the anchor and the samples different from the anchor, 
  and minimizing the distance between features of the anchor and the augmentation.
  The calculation of SelfCL is shown in Formula 1 and 2.
  Formula 1 indicates that the total SelfCL of a batch is the sum of the loss of each sample.
  Formula 2 shows how to calculate the loss of each sample.
  The $z_i$, $z_{j(i)}$, $z_k$
  refers to the features extracted from the anchor, the positive, and the negative.
  The value of $\mathbb{L}^{self}_i$  represents 
  the negative logarithm of 
  the proportion of Euclidean Distances \cite{danielsson1980euclidean} of
  $z_i$ and $z_{j(i)}$ to the sum of Euclidean Distances of $z_i$ and all $z_k$.
  Decreasing the value of $\mathbb{L}^{self}_i$ increases the proportion of Euclidean Distance of 
  $z_i$ and $z_{j(i)}$ to the sum of Euclidean Distances of $z_i$ and all $z_k$, 
  which makes the feature of the anchor similar to the feature of the augmentation, 
  and makes the feature of the anchor different from all features of other samples.
  The SelfCL greatly promotes the development of contrastive learning,
  however, the SelfCL doesn't introduce the label and there is only one positive
  by default. The SelfCL is unavailable in supervised learning,
  and cannot be utilized by the steganalysis schemes.

\begin{equation}
\mathbb{L}^{sup}=\sum^{2N}_{i=1}\mathbb{L}^{sup}_{i}
\end{equation}
\begin{equation}
	\mathbb{L}^{sup}_{i}
	=
	\frac{-1}{2N_{\widetilde{y}_{i}} -1}
	\sum^{2N}_{j=1}
	\ell_{i \neq j}
	\cdot
	\ell_{\widetilde{y}_{i}=\widetilde{y}_{j}}
	\cdot
	\log{\frac{\exp(z_{i} \cdot z_{j} / \tau)}
		{\sum^{2N}_{k=1}\ell_{i \neq k} \cdot \exp(z_{i} \cdot z_{k} / \tau)}
	}
\end{equation}

  Recently, Google researchers propose the Supervised Contrastive Loss (SupCL) 
  to introduce contrastive learning into supervised learning in NeurIPS 2020 \cite{khosla2020supervised}.
  The SupCL takes the labels of samples into account and contrasts the anchor with more than one positive.
  With the help of the labels, the contrast occurs between the anchor, 
  the samples whose labels are the same as that of the anchor (the positive),
  and the samples whose labels are different from that of the anchor (the negative).
  The SupCL tries to bring the features of samples of the same category close, 
  and keep the features of samples of different categories far away.
  The calculation of SupCL is shown in Formula 3 and 4.
  Compared to the $\mathbb{L}^{self}_i$ shown in Formula 2,
  the $\mathbb{L}^{sup}_i$ 
  can be understood as the normalized sum of
  $(2N_{\widetilde{y}_{i}}-1)$\footnote{ $2N_{\widetilde{y}_{i}}$  is the number of samples that is similar to the anchor in the batch}
  times
  $\mathbb{L}^{self}_i$ computing,
  and each $\mathbb{L}^{self}_i$
  makes the anchor similar to the only one positive.
  By
  $(2N_{\widetilde{y}_{i}}-1)$
  times $\mathbb{L}^{self}_i$ computing,
  the anchor is similar to all samples of the same category.
\begin{figure}[htbp]
  \centering
  \subfigure[$4\times3$ times computing]{
    \includegraphics[width=0.2\linewidth]{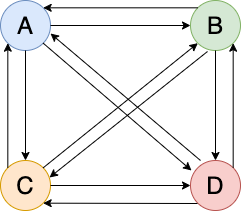}
  }
  \qquad
  \qquad
  \subfigure[$4\times3\div2$ times computing]{
    \includegraphics[width=0.2\linewidth]{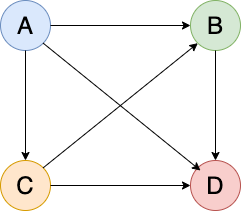}
  }
  \qquad
  \qquad
  \subfigure[$3$ times computing]{
    \includegraphics[width=0.2\linewidth]{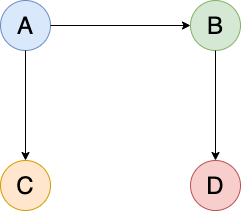}
  }
  \caption{The three schemes to make samples (A, B, C, and D) of the same category similar.
   The black arrow represents once $\mathbb{L}^{self}_i$ computing.}
  \label{fig:supcl}
\end{figure}
  There is redundant computing in the SupCL.
  To make it easier for readers to understand, Figure~\ref{fig:supcl} is drawn to 
  explain the redundant computing in the SupCL.
  Three schemes make $N=4$ samples of the same category similar.
  Figure~\ref{fig:supcl} (a)
  represents the scheme used by the SupCL, which has 
  $N\times(N-1)=12$  times
  $\mathbb{L}^{self}_i$ computing.
  Figure~\ref{fig:supcl} (b)
  represents the scheme that removes the repeat computing 
  ($A \rightarrow B$ is repeated to $B \rightarrow A$ ), 
  and there are $N\times(N-1)\div2=6$ times computing.
  The scheme in Figure~\ref{fig:supcl} (c) 
  considers the transitivity of similarity(
  $A \rightarrow B$ and $B \rightarrow C$,
  which makes $A \rightarrow C$ unnecessary),
  and there is $N-1=3$ times computing left.
  All schemes in Figure~\ref{fig:supcl} can make samples of the same category similar.
  However, the scheme used by the SupCL needs more computing, which means that 
  the contrastive loss in supervised learning has redundant computing.

\section{Proposed methods}

  Based on the analysis above, we propose the Steganalysis Contrastive Framework (SCF) that 
  introduces contrastive learning into steganalysis for the first time. 
  To eliminate the redundant computing in the existing contrastive loss,
  a novel Steganalysis Contrastive Loss (StegCL) is 
  designed based on the equivalence and transitivity of similarity.

\subsection{Steganalysis Contrastive Framework}

\begin{figure}[htbp]
  \centering
  \includegraphics[width=0.99\linewidth]{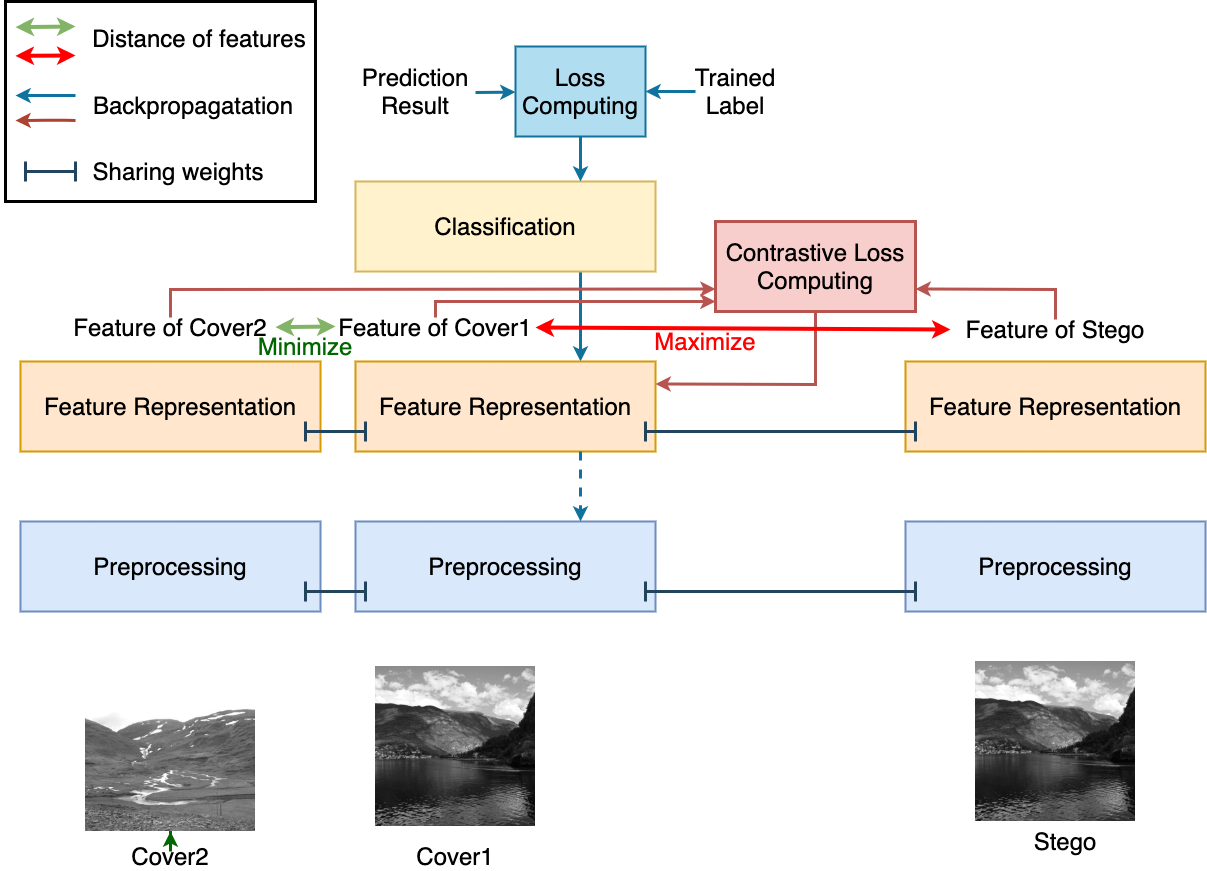}
  \caption{The architecture of the SCF.}
  \label{fig:4}
\end{figure}

  The architecture of the Steganalysis Contrastive Framework (SCF) is shown in Figure~\ref{fig:4}.
  The main difference between the SCF and the basic framework of existing steganalysis DNNs (seen in Figure~\ref{fig:2})
  is that the
  SCF adds a contrastive loss computing module. This module computes the contrastive loss 
  with the features extracted by the feature representation module.
  The SCF retains the backpropagation from the loss computed with the prediction result and the trained label, 
  which maximizes the distance between features of cover and stego.
  Meanwhile, the SCF introduces a new backpropagation from the contrastive loss, 
  which not only maximizes the distance between features of cover and stego
  but also minimizes the distance between features of samples of the same category.

\subsection{Steganalysis Contrastive Loss}

\begin{equation}
\mathbb{L}^{steg}=\sum^{2N}_{i=1}\mathbb{L}^{steg}_{i}
\end{equation}

\begin{equation}
	\mathbb{L}^{steg}_{i}=-\log{
		\frac{\exp({z_i} \cdot z_{j(Random \cdot \ell_{(i) \neq (j)} \cdot \ell_{ \widetilde{y}_i=\widetilde{y}_{j}})} / \tau)}
		{\sum^{2N}_{k=1}\ell_{i \neq k} \cdot \ell_ {\widetilde{y}_i\neq\widetilde{y}_j}
				\cdot \exp(z_i \cdot z_k / \tau)}
	}
\end{equation}

\begin{figure}[H]
  \centering
  \subfigure[Initialization]{
    \includegraphics[width=0.19\linewidth]{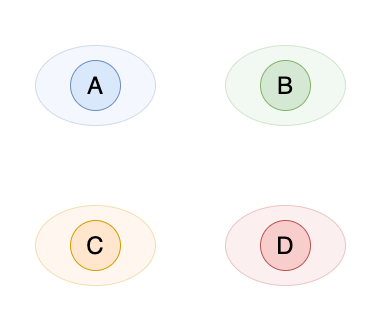}
  }
  \quad
  \subfigure[$A \rightarrow D$]{
    \includegraphics[width=0.19\linewidth]{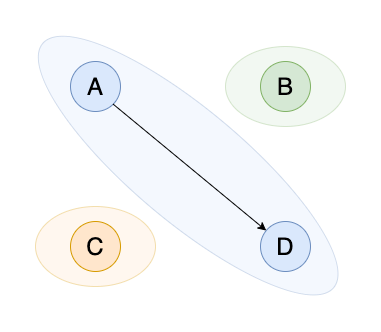}
  }
  \quad
  \subfigure[$B \rightarrow C$]{
    \includegraphics[width=0.19\linewidth]{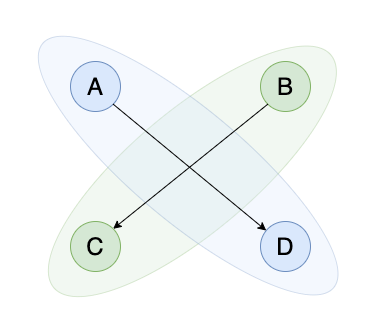}
  }
  \quad
  \subfigure[$C \rightarrow D$]{
    \includegraphics[width=0.19\linewidth]{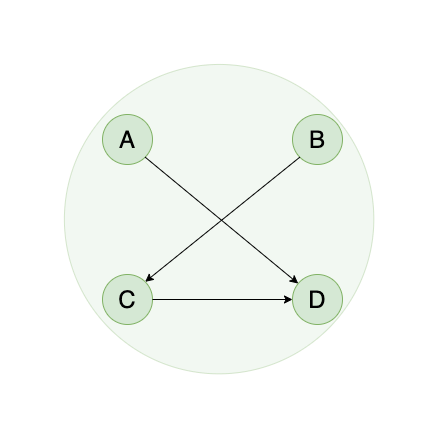}
  }
  \caption{The figures explain how Algorithm\ref{algo} works.
  The A, B, C, and D in the circle are the samples of the same category. The ellipses are the sets that contain the same category
  samples. The arrow means once $\mathbb{L}^{self}_i$ computing.} 
  \label{fig:algoin4}
\end{figure}

  The Steganalysis Contrastive Loss (StegCL) is shown in Formula 5 and 6.
  Compared to the $\mathbb{L}^{sup}_i$ shown in Formula 4, the $\mathbb{L}^{steg}_i$ only contains once $\mathbb{L}^{self}_i$ computing.
  The advantage of the StegCL is that it minimizes the features of samples of the same category without redundant computing.
  The core of the StegCL is the Random Selection Strategy (RSS) that is the 
  $(Random \cdot \ell_{(i) \neq (j)} \cdot \ell_{ \widetilde{y}_i=\widetilde{y}_{j}})$ shown in Formula 6.
  The RSS selects the same category sample in different set of the anchor as the positive.
  Algorithm~\ref{algo} presents how the RSS works.
  To make it easier for readers to understand,
  Figure~\ref{fig:algoin4}
  is drawn to explain how Algorithm~\ref{algo} executes when there are 4 samples of the same category.
  Figure~\ref{fig:algoin4} (a) shows that samples are not similar to each other at the start.
  Figure~\ref{fig:algoin4} (b) shows that anchor A select sample D as the positive, which makes A similar to D and merges the sets.
  Figure~\ref{fig:algoin4} (c) shows that anchor B select sample C as the positive, which makes B similar to C and merges the sets.
  Figure~\ref{fig:algoin4} (d) shows that anchor C select sample D as the positive, which makes the A, B, C and D are similar to each other and merges the sets.
  With the help of the RSS used in StegCL, each anchor only need to be similar to one 
  positive, and the samples of the same category are similar to each other finally.

\IncMargin{2em}
\begin{algorithm}
  \SetKwData{Set}{set}\SetKwData{Union}{U}
  \SetKwData{Image}{image}
  \SetKwFunction{SetFun}{SetFunction}
  \SetKwFunction{GetElementSet}{GetElementSet}
  \SetKwFunction{RandomChoseFromSet}{RandomSelectFromSet}
  \SetKwFunction{UnionFunction}{UnionSetFunction}
  \SetKwFunction{MakeClose}{MakeClose}
  \SetKwInOut{Input}{input}\SetKwInOut{Output}{output}

  \Input{A batch $Im$ of $N$ number}
  \Output{None}
  \BlankLine

  \Union $\leftarrow$ \SetFun{$Im[1:N]$}\;
  \For{$i\leftarrow 1$ \KwTo $N$}{
    \Set $\leftarrow$ \SetFun{$Im[i]$}\;
  }

  \For{$i\leftarrow 1$ \KwTo $N-1$}{
    {\color{red}
    \Set $\leftarrow$ \GetElementSet{$Im[i]$}\;
    \Set $\leftarrow$ \Union $-$ \Set\;
    \Image $\leftarrow$ \RandomChoseFromSet{\Set}\;
    }
    \MakeClose{$Im[i]$ and \Image}\;
    \UnionFunction{$Im[i]$ and \Image}\;
  }

  \caption{The algorithm that utilizes RSS to make samples of the same category similar.
  The RSStrategy is the red part.}
  \label{algo}
\end{algorithm}
\DecMargin{2em}

\section{Experiments}

  To analyze the performance of the SCF and the StegCL, 
  the detection accuracy, generalization, and computing complexity are evaluated.
  To evaluate the detection accuracy,
  the DNNs optimized with the SCF are tested on WOW \cite{holub2012designing},
  S-UNIWARD \cite{holub2014universal},
  and MiPOD \cite{sedighi2015content}
  at $0.2$ bpp and $0.4$ bpp payloads, 
  and the results are compared to that of the original DNNs without the optimization of the SCF.
  To evaluate the generalization,
  the detection accuracy of the optimized DNNs is compared to
  that of the original DNNs in case of mismatch.
  To evaluate the computing complexity,
  the StegCL and the SupCL are used by the SCF, and the training time is compared.
  The details of the experiment are as follows.

\subsection{Setups}
\subsubsection{Dataset}

  The dataset used in experiments is BOSSbase $1.01$ \cite{bas2011break} which contains $10,000$ Portable GrayMap format images.
  These images are $512\times512$ pixels.
  Considering that the GPU used in experiments is GeForce $2080$Ti that has merely $12$G memory,
  these images are resized to $256\times256$ pixels by the {\it imresize} function in MATLAB with default settings.
  For simplification, the dataset with resized images is called BOSSbase256.
  The images in BOSSbase256 are used as covers in experiments.
  To generate the stegoes from the covers, four self-adaptive steganography algorithms, HUGO \cite{pevny2010using},
  WOW \cite{holub2012designing},
  S-UNIWARD \cite{holub2014universal},
  MiPOD \cite{sedighi2015content} are facilitated.
  There are $0.1$ bpp, $0.2$ bpp, $0.3$ bpp, and $0.4$ bpp four payloads in experiments.
  S-UNIWARD $0.4$ bpp BOSSbase256 represents the dataset that contains $10,000$ covers and $10,000$ stegoes generated
  by S-UNIWARD \cite{holub2014universal} at $0.4$ bpp payload.
  Each dataset is randomly divided into three parts in a ratio of $6$:$1$:$3$ for training, validation, and testing.

\subsubsection{Metric}

  The performance of detection accuracy is measured with the total classification error probability 
  on the test subset of the dataset 
  under equal priors $P_E$$=\min_{P_{FA}}\frac{1}{2}(P_{FA}+P_{MD})$
  \cite{boroumand2018deep, you2020siamese, yedroudj2018yedroudj, ye2017deep},
  where $P_{FA}$ and $P_{MD}$ are the false-alarm and missed-detection probabilities.

\subsection{Experimental designs}

      \subsubsection{Evaluation of detection accuracy}

      The $P_E$ of original DNNs and DNNs optimized with the SCF are compared on different datasets.
         With WOW \cite{holub2012designing}, S-UNIWARD \cite{holub2014universal}, and MiPOD \cite{sedighi2015content} at $0.2$ bpp and $0.4$ bpp,
      six different datasets are constructed.
      YeNet \cite{ye2017deep}, YedroudjNet \cite{yedroudj2018yedroudj}, KeNet \cite{you2020siamese}, 
      CL\footnote{The CL suffix means the DNNs that optimized with the StegCL by the SCF}-YeNet,
      CL-YedroudjNet, and CL-KeNet are trained and tested on these datasets.
      The $P_E$ of the DNNs are compared.

      \subsubsection{Evaluation of generalization}

        \subsubsection*{Steganography Algorithm Mismatch}
         
          The DNNs are trained on algorithm known datasets and tested on algorithm unknown datasets.
          HUGO \cite{pevny2010using}, WOW \cite{holub2012designing}, S-UNIWARD \cite{holub2014universal}, MiPOD \cite{sedighi2015content}, $0.4$ bpp BOSSbase256 are
      constructed. KeNet \cite{you2020siamese} and CL-KeNet are trained on one of these datasets and tested on
      all four datasets. Meanwhile, the $P_E$ of KeNet and CL-KeNet are compared with each other.

    \subsubsection*{Payload Mismatch}

          The DNNs are trained on payload known datasets and tested on payload unknown datasets.
          S-UNIWARD \cite{holub2014universal} $0.1$ bpp, $0.2$ bpp, $0.3$ bpp, $0.4$ bpp BOSSbase256 are
          constructed. KeNet \cite{you2020siamese} and CL-KeNet are trained on one of these datasets and tested on
      all four datasets. Meanwhile, the $P_E$ of KeNet and CL-KeNet are compared with each other.

      \subsubsection{ Comparation of computing complexity between StegCL and SupCL \cite{khosla2020supervised}}

      The
      SupCL-KeNet (KeNet+SCF+SupCL \cite{khosla2020supervised}),
      KeNet \cite{you2020siamese}, and StegCL-KeNet (KeNet+SCF+StegCL)
      are trained and tested on S-UNIWARD $0.4$ bpp BOSSbase256.
      The training time and the $P_E$ of the DNNs are compared.


\subsection{Experimental results}

\subsubsection{Evaluation of detection accuracy}

\begin{figure}[htbp]
  \centering
  \subfigure[0.2 bpp]{
    \includegraphics[width=0.46\linewidth]{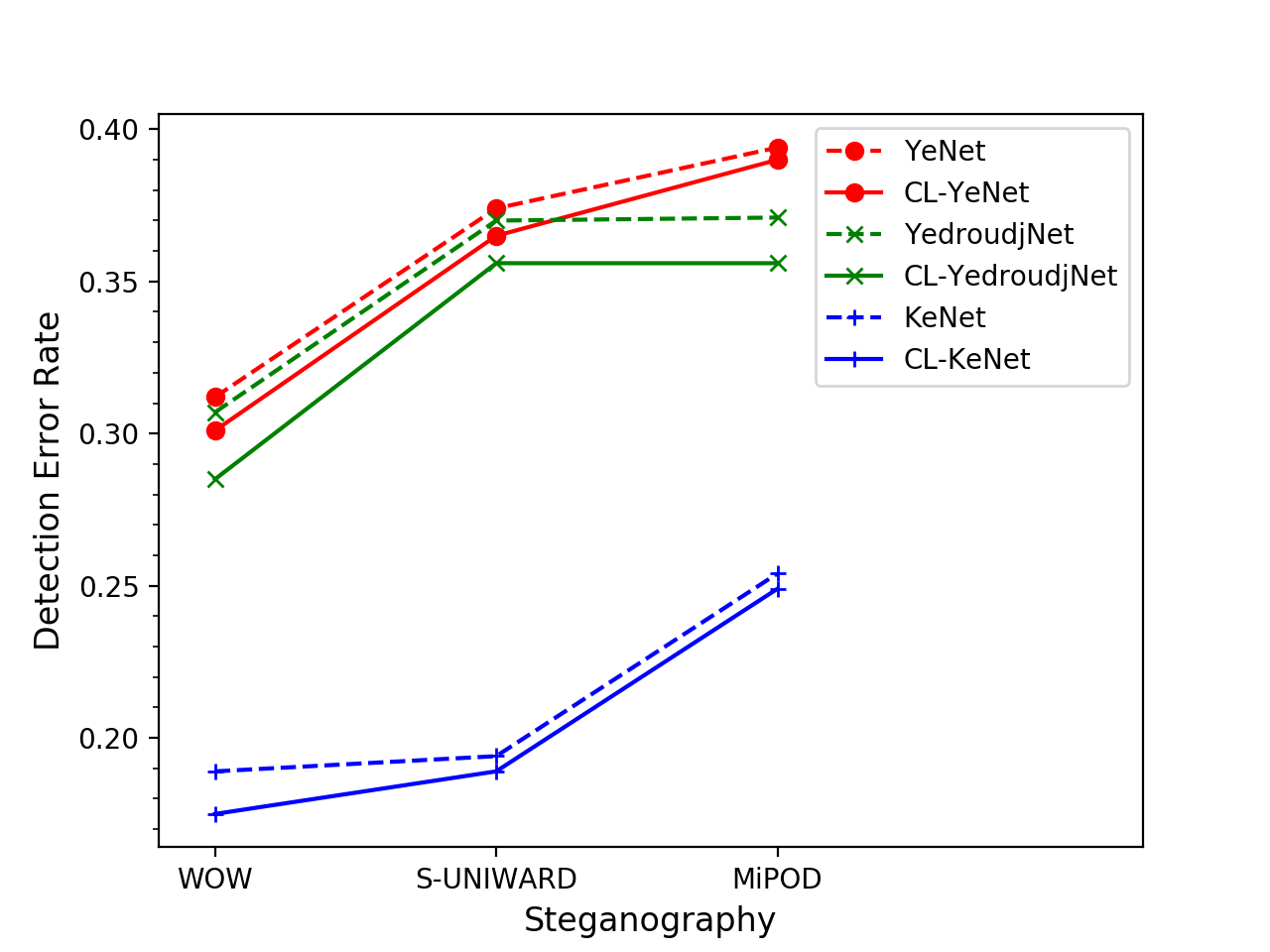}
  }
  \subfigure[0.4 bpp]{
    \includegraphics[width=0.46\linewidth]{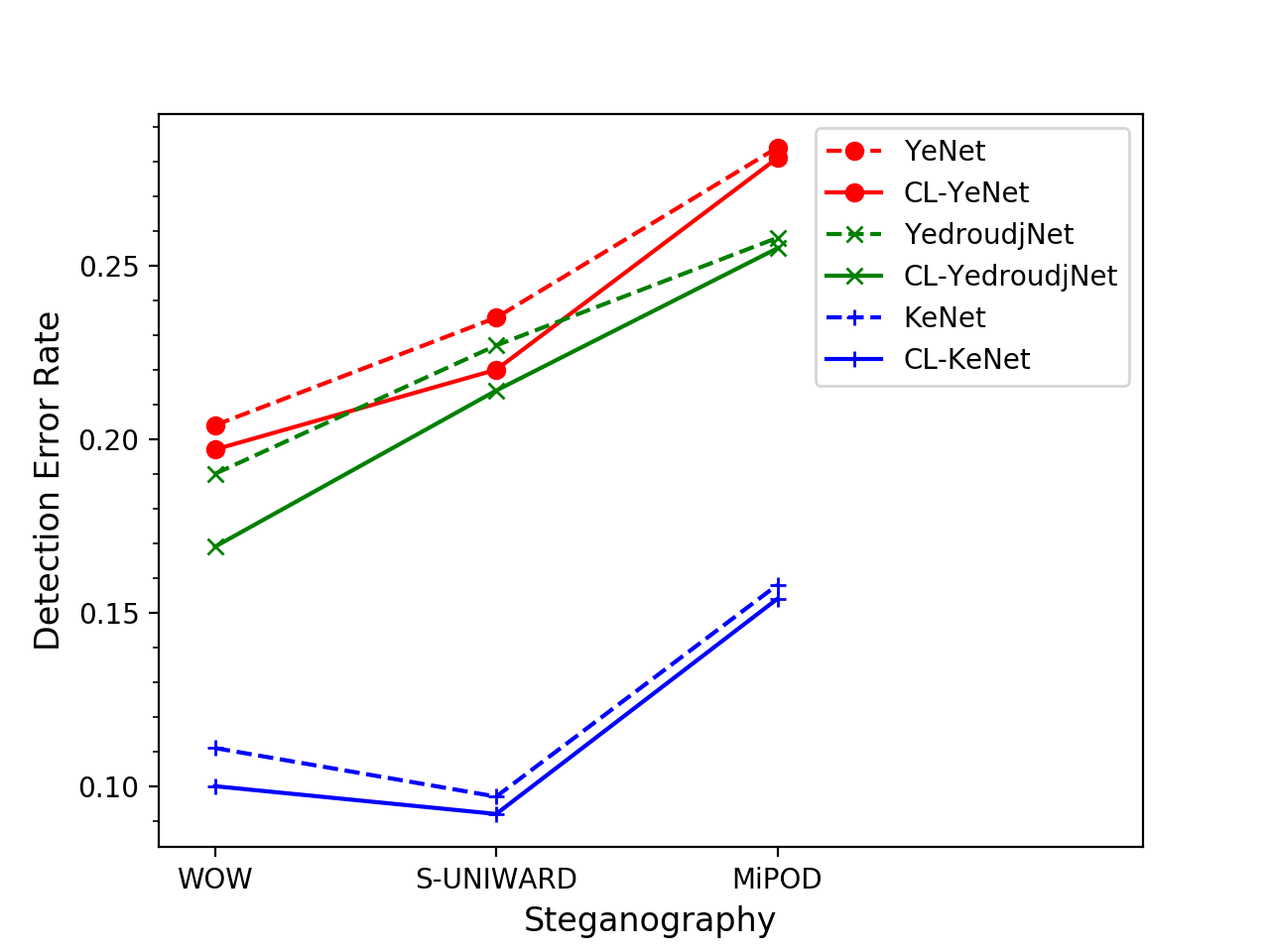}
  }
  \caption{The line graphs that represent the $\boldsymbol{P_E}$ of YeNet \cite{ye2017deep}, CL-YeNet, YedroudjNet \cite{yedroudj2018yedroudj}, CL-YedroudjNet, KeNet \cite{you2020siamese}, and CL-KeNet on
  WOW \cite{holub2012designing}, S-UNIWARD \cite{holub2014universal}, MiPOD \cite{sedighi2015content} at 0.2 bpp and 0.4 bpp.}
  \label{fig:lines}
\end{figure}

  The line graphs of $P_E$ are shown in Figure~\ref{fig:lines}.
  The horizontal axis represents steganography algorithms, and the vertical axis indicates the detection error rates $P_E$.
  The dashed lines represent the $P_E$ of YeNet \cite{ye2017deep}, YedroudjNet \cite{yedroudj2018yedroudj}, and KeNet \cite{you2020siamese}.
  The solid lines represent the $P_E$ of CL-YeNet, CL-KeNet, and CL-YedroudjNet.
  The solid lines are lower than the dashed lines with the same color, which
  means that the detection accuracy is improved with the help of the SCF.
  The detailed experimental results are shown in Table~\ref{tab:I}.
  The $P_E$ of the DNNs optimized with the SCF is $0.5\%$ to $3\%$ lower than that of the original DNNs,
  which means the SCF increases the detection accuracy of steganalysis.

\begin{table}
  \small {
    \caption{The $\boldsymbol{P_E}$ of YeNet \cite{ye2017deep}, CL-YeNet, YedroudjNet \cite{yedroudj2018yedroudj}, CL-YedroudjNet, KeNet \cite{you2020siamese}, and CL-KeNet on
  WOW \cite{holub2012designing}, S-UNIWARD \cite{holub2014universal}, MiPOD \cite{sedighi2015content} at 0.2 bpp and 0.4 bpp.}
  \label{tab:I}
  \resizebox{.99\linewidth}{!}{
  \begin{tabular}{ccccccc}
    \toprule
    {\bfseries Steganography}	& \multicolumn{2}{c}{WOW \cite{holub2012designing}} &
    \multicolumn{2}{c}{S-UNIWARD \cite{holub2014universal}} &
    \multicolumn{2}{c}{MiPOD \cite{sedighi2015content}} \\
    {\bfseries Payload}	          &  0.2 bpp&	0.4 bpp&	  0.2 bpp&	0.4 bpp&	0.2 bpp  &0.4 bpp \\
    \midrule
    YeNet \cite{ye2017deep}	            &           .312	&           .204	&             .374	&           .235	&           .394	&          .284 \\
    CL-YeNet	    & \bfseries .301	&\bfseries  .197	&\bfseries    .365	&\bfseries  .220	&\bfseries  .390	&\bfseries .281 \\
    \midrule
    YedroudjNet \cite{yedroudj2018yedroudj}	      &           .307	&           .190	&             .370	&           .227	&           .371	&          .258 \\
    CL-YedroudjNet&	\bfseries .285 &\bfseries  .169	&\bfseries    .356	&\bfseries  .214	&\bfseries  .356	&\bfseries .255 \\
    \midrule
    KeNet \cite{you2020siamese}	            &           .189	&           .111	&             .194	&           .097	&           .254	&          .158 \\
    CL-KeNet	    & \bfseries .175	&\bfseries  .100	&\bfseries    .189	&\bfseries  .092	&\bfseries  .249	&\bfseries .154 \\
  \bottomrule
  \end{tabular}
}
}
\end{table}

\begin{figure}[htbp]
  \centering
  \subfigure[from KeNet \cite{you2020siamese}]{
    \includegraphics[width=0.25\linewidth]{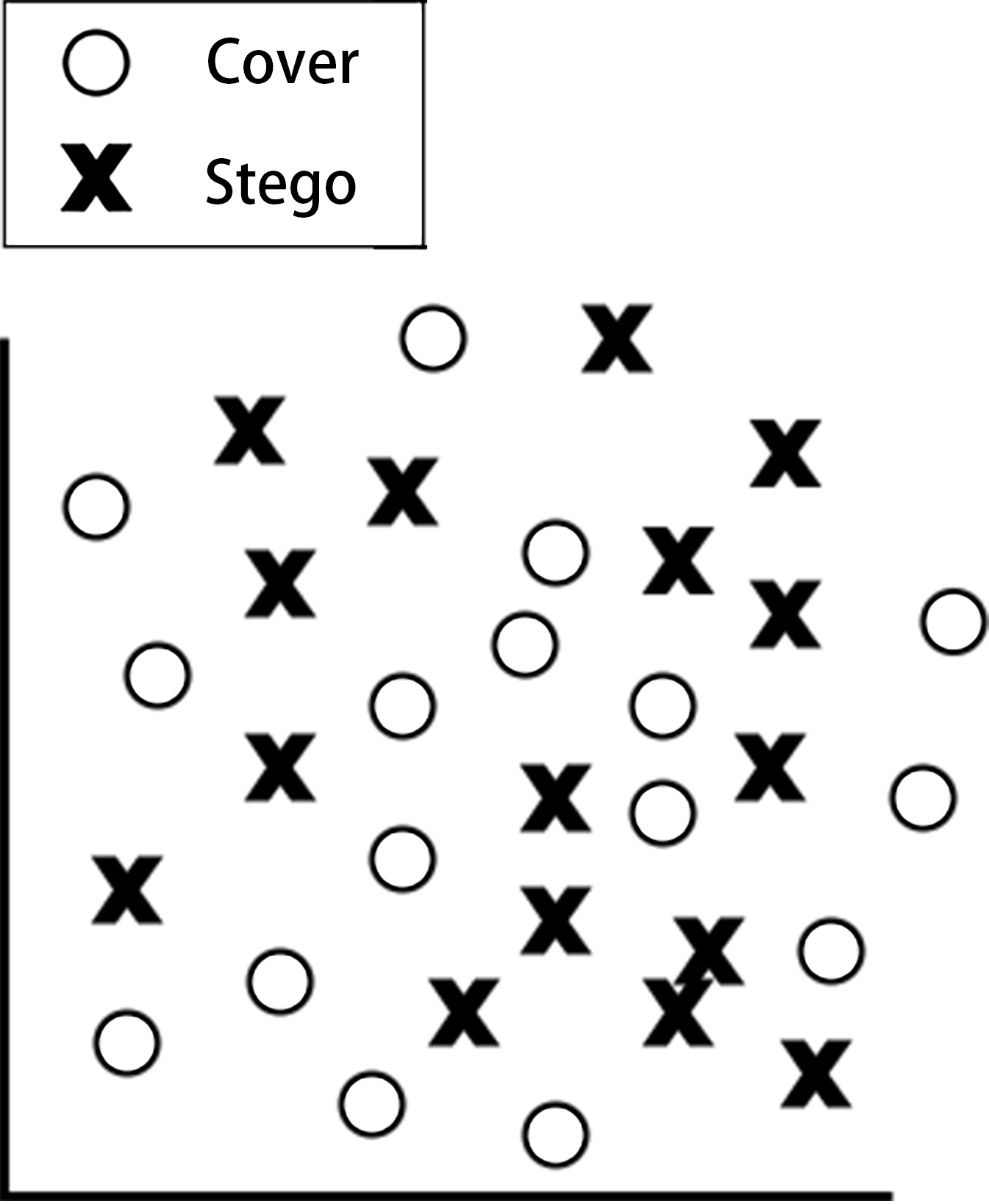}
  }
  \quad
  \quad
  \quad
  \quad
  \quad
  \subfigure[from CL-KeNet]{
    \includegraphics[width=0.25\linewidth]{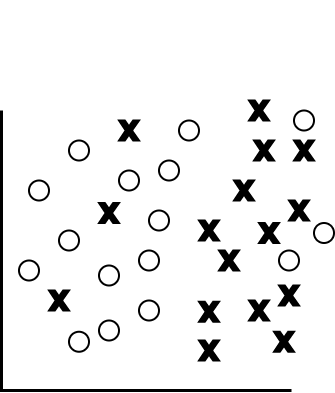}
  }
  \caption{The 2-D t-SNE features}
  \label{fig:features}
\end{figure}

  The Feature Representation Modules (FRMs) of KeNet \cite{you2020siamese} and CL-KeNet are utilized to extract the features.
  Two batch features are extracted from the same batch samples by the FRMs.
  Then the features are reduced to $2$-Dimensions with t-SNE \cite{maaten2008visualizing}.
  These $2$-D features are plotted in two axes in Figure~\ref{fig:features}.
  The features in Figure~\ref{fig:features} (b) are more discriminating than those in Figure~\ref{fig:features} (a),
  which means the SCF minimizes the distance between features of samples of the same category.

\subsubsection{Evaluation of generalization}

\begin{table}
  \small {
    \caption{The $\boldsymbol{P_E}$ of KeNet \cite{you2020siamese} and CL-KeNet in case of steganography algorithm mismatch under the same 0.4 bpp payload.}
  \label{tab:II}
  \resizebox{1.\linewidth}{!}
  {
  \begin{tabular}{cccccc}
    \toprule
    \bfseries MODELS & {\bfseries	TRA $\backslash$ TES} & 	WOW \cite{holub2012designing} &
    HUGO \cite{pevny2010using} &S-UNI* \cite{holub2014universal} & MiPOD \cite{sedighi2015content} \\
    \midrule
    KeNet \cite{you2020siamese}	      &  \multirow{2}*{WOW \cite{holub2012designing}}	      &      .110     &	     .383 &   	  .205 &   	.289 \\
    CL-KeNet&	                          & \bf  .100     &	\bf .378 &\bf	  .191 &\bf	.287 \\
    \midrule
    KeNet	      &  \multirow{2}*{HUGO \cite{pevny2010using}}	    &      .318     &	     .174 &   	  .321 &   	.320 \\
    CL-KeNet&	                          & \bf  .302     &	\bf .153 &\bf	  .311 &\bf	.308 \\
    \midrule
    KeNet	      &  \multirow{2}*{S-UNI* \cite{holub2014universal}} &	     .106     &      .276 &	     .097 &     .214 \\
    CL-KeNet&	                          & \bf  .101     &	\bf  .268 &\bf	  .092 &\bf	.211 \\
    \midrule
    KeNet	      &  \multirow{2}*{MiPOD \cite{sedighi2015content}}	    &      .150     &	     .262 &   	  .167 &   	.158 \\
    CL-KeNet&                           & \bf  .149 &	\bf  .250&\bf	  .154 &\bf	.153\\
  \bottomrule
  \end{tabular}
  }
}
\end{table}

\begin{table}
  \small {
    \caption{The $\boldsymbol{P_E}$ of KeNet \cite{you2020siamese} and CL-KeNet in case of payload mismatch under the same S-UNIWARD \cite{holub2014universal} steganography algorithm.}
  \label{tab:III}
  \resizebox{.85\linewidth}{!}{
  \begin{tabular}{cccccc}
    \toprule
    \bfseries MODELS & {\bfseries	TRA $\backslash$ TES} & 0.1 bpp & 0.2 bpp &0.3 bpp & 0.4 bpp \\
    \midrule
    KeNet \cite{you2020siamese}	      &  \multirow{2}*{0.1 bpp}	      &      .382    &	     .254 &	     .215 &   	.174 \\
    CL-KeNet&	                              &\bf   .375     &	\bf .248   &	 \bf  .212 &\bf .153 \\
    \midrule
    KeNet	      &  \multirow{2}*{0.2 bpp}	      &      .395     &	     .238 &	    .160 &   	.158 \\
    CL-KeNet&	                              &\bf   .393     &	 \bf .230 &	 \bf .148 &\bf	.153 \\
    \midrule
    KeNet	      &  \multirow{2}*{0.3 bpp}        &   	 .400     &      .281 &	     .153 &   	.110 \\
    CL-KeNet&	                              &\bf   .399     &	 \bf .279 &	 \bf .139 &\bf	.100 \\
    \midrule
    KeNet	      &  \multirow{2}*{0.4 bpp}	      &      .442     &	     .302 &	     .203 &   	.097 \\
    CL-KeNet&                               &\bf   .431     &	 \bf .298 &	 \bf .200 &\bf	.092 \\
  \bottomrule
  \end{tabular}
  }
}
\end{table}

  The $P_E$ of KeNet \cite{you2020siamese} and CL-KeNet in case of mismatch is shown in Table~\ref{tab:II} and Table~\ref{tab:III}.
  Table~\ref{tab:II} shows the $P_E$ in case of steganography algorithm mismatch.
  When the DNN is trained on one steganography and tested on HUGO \cite{pevny2010using}, WOW \cite{holub2012designing},
  S-UNIWARD \cite{holub2014universal} and MiPOD \cite{sedighi2015content},
  the $P_E$ of CL-KeNet are all lower than that of KeNet \cite{you2020siamese}, which represents that the SCF improves
  the generalization of steganalysis DNNs when steganography algorithm mismatch happens.
  Table~\ref{tab:III} shows the $P_E$ in case of payload mismatch.
  The $P_E$ of CL-KeNet are whole lower than that of KeNet \cite{you2020siamese} when the DNN is trained on one payload
  and tested on four payloads, which means that the SCF improves the generalization
  of steganalysis DNNs when payload mismatch happens.

\subsubsection{Comparation of computing complexity between StegCL and SupCL \cite{khosla2020supervised}}

\begin{table}
  \small {
    \caption{The Training time of each epoch and $\boldsymbol{P_E}$ of KeNet \cite{you2020siamese}, SupCL-KeNet, and StegCL-KeNet 
  on the S-UNIWARD 0.4 bpp BOSSbase256.}
  \label{tab:IV}
  \resizebox{0.75\linewidth}{!}{
    \begin{tabular}{ccc}
    \toprule
      \bfseries MODELS	& \bf Training time of each epoch &	$\boldsymbol{P_E}$ \\
    \midrule
      KeNet \cite{you2020siamese}	& {\bfseries 170 seconds}	& .0979 \\
      SupCL \cite{khosla2020supervised}-KeNet	& 2100 seconds & .0926 \\
    StegCL-KeNet & {\bfseries 200 seconds} &	{ .0921} \\
  \bottomrule
  \end{tabular}
  }
}
\end{table}

  Table~\ref{tab:IV} shows the training time and  $P_E$ of KeNet \cite{you2020siamese}, SupCL-KeNet, and StegCL-KeNet.
  In the column of training time of each epoch, the values of KeNet \cite{you2020siamese} and StegCL-KeNet are close, but the value of
  SupCL-KeNet is far from that of these.
  The $P_E$ of StegCL-KeNet is smaller than that of SupCL-KeNet.
  The experimental result shows that the designed StegCL costs less training time than that of the contrastive loss in supervised learning,
  without decreasing the detection accuracy.

\section{Conclusion}
  This paper introduces contrastive learning into steganalysis for the first time and 
  proposes the Steganalysis Contrastive Framework (SCF).
  The SCF is utilized to optimize well-known steganalysis DNNs, YeNet, YedroudjNet, and KeNet.
  The experimental results indicate that contrastive learning can greatly improve 
  the detection accuracy and generalization of existing steganalysis DNNs.
  Meanwhile, we optimize the contrastive loss in supervised learning and design a novel StegCL based on 
  the equivalence and transitivity of similarity.
  The StegCL eliminates the redundant computing in the contrastive loss in supervised learning and 
  reduces the computing complexity.
  The future work will further follow the idea of contrastive learning, mining
  the intrinsic correlation of cover and stego to improve the performance
  of the steganalysis schemes.


\bibliographystyle{ACM-Reference-Format}
\bibliography{sample-base}

%
%
%
%
%
%
%
%

\end{document}